# Comparative performance of ensemble models in predicting dental provider types: insights from fee-for-service data

## Desempeño Comparativo de Modelos de Conjunto en la Predicción de Tipos de Proveedores Dentales: Perspectivas a partir de Datos de Pago por Servicio

Mohammad Subhi Al-Batah[1] , Muhyeeddin Alqaraleh[2] , Mowafaq Salem Alzboon[1] , Abdullah Alourani[3] 

[1]Jadara University, Faculty of Information Technology. Irbid, Jordan.
[2]Zarqa University, Faculty of Information Technology. Zarqa, Jordan.
[3]Qassim University, Department of Management Information Systems, College of Business and Economics, Buraydah 51452, Saudi Arabia.

**ABSTRACT**

Dental provider classification plays a crucial role in optimizing healthcare resource allocation and policy planning. Effective categorization of providers, such as standard rendering providers and safety net clinic (SNC) providers, enhances service delivery to underserved populations. To evaluate the performance of machine learning models in classifying dental providers using a 2018 dataset. A dataset of 24,300 instances with 20 features was analyzed, including beneficiary and service counts across fee-for-service (FFS), Geographic Managed Care, and Pre-Paid Health Plans. Providers were categorized by delivery system and patient age groups (0-20 and 21+). Despite 38,1 % missing data, multiple machine learning algorithms were tested, including k-Nearest Neighbors (kNN), Decision Trees, Support Vector Machines (SVM), Stochastic Gradient Descent (SGD), Random Forest, Neural Networks, and Gradient Boosting. A 10-fold cross-validation approach was applied, and models were evaluated using AUC, classification accuracy (CA), F1-score, precision, and recall. Neural Networks achieved the highest AUC (0,975) and CA (94,1 %), followed by Random Forest (AUC: 0,948, CA: 93,0 %). These models effectively handled imbalanced data and complex feature interactions, outperforming traditional classifiers like Logistic Regression and SVM. Advanced machine learning techniques, particularly ensemble and deep learning models, significantly enhance dental workforce classification. Their integration into healthcare analytics can improve provider identification and resource distribution, benefiting underserved populations.

**RESUMEN**

La clasificación de proveedores dentales desempeña un papel crucial en la optimización de la asignación de recursos de atención médica y la planificación de políticas. La categorización efectiva de proveedores, como proveedores de prestación estándar y proveedores de clínicas de red de seguridad (SNC), mejora la prestación de servicios a las poblaciones desatendidas. Para evaluar el rendimiento de los modelos de aprendizaje automático en la clasificación de proveedores dentales utilizando un conjunto de datos de 2018, se analizó un conjunto de datos de 24 300 instancias con 20 características, incluyendo el recuento de beneficiarios y servicios en pagos por servicio (FFS), Atención Administrada Geográficamente y Planes de Salud Prepagos. Los proveedores se categorizaron por sistema de prestación y grupos de edad de pacientes (0-20 y 21+). A pesar del 38,1 % de datos faltantes, se probaron múltiples algoritmos de aprendizaje automático, incluyendo k-Vecinos Más Cercanos (kNN), Árboles de Decisión, Máquinas de Vectores de Soporte (SVM), Descenso de Gradiente Estocástico (SGD), Bosque Aleatorio, Redes Neuronales y Potenciación de Gradiente. Se aplicó un enfoque de validación cruzada de 10 pasos y los modelos se evaluaron mediante el AUC, la precisión de clasificación (CA), la puntuación F1, la precisión y la recuperación. Las redes neuronales obtuvieron los valores más altos de AUC (0,975) y CA (94,1 %), seguidas de Random Forest (AUC: 0,948, CA: 93,0 %). Estos modelos gestionaron eficazmente datos desequilibrados e interacciones complejas de características, superando a clasificadores tradicionales como la regresión logística y la SVM. Las técnicas avanzadas de aprendizaje automático, en particular los modelos de aprendizaje conjunto y profundo, mejoran significativamente la clasificación del personal odontológico. Su integración en el análisis de datos de atención médica puede mejorar la identificación de proveedores y la distribución de recursos, beneficiando a las poblaciones desatendidas.

**Palabras clave:** Aprendizaje Automático; Clasificación de Proveedores Dentales; Modelos Conjuntos; Analítica Sanitaria; Clínicas de Red de Seguridad.

## INTRODUCTION

The classification of healthcare providers is pivotal for optimizing resource allocation, improving service delivery, and addressing disparities in access to care. In dentistry, distinguishing between standard rendering providers and safety net clinics (SNCs) is particularly crucial, as SNCs predominantly serve low-income, uninsured, and vulnerable populations.[1] Accurate classification enables policymakers to identify gaps in service coverage, allocate funding effectively, and design targeted interventions. However, traditional statistical methods often struggle with high-dimensional, imbalanced datasets common in healthcare analytics.[2] This gap has spurred interest in machine learning (ML), which offers advanced tools for pattern recognition, predictive modeling, and decision-making under uncertainty.[3]

Recent advancements in ML have revolutionized healthcare applications, from disease diagnosis to workforce planning. For instance, ensemble models like Random Forest and Gradient Boosting excel in handling heterogeneous data and mitigating overfitting, while neural networks leverage deep architectures to capture complex interactions. In dental research, ML has been applied to tasks such as caries detection, implant classification, and age estimation. However, few studies focus on provider classification, despite its implications for equitable care delivery. Existing literature emphasizes technical challenges, including missing data, class imbalance, and feature relevance—issues that directly impact model generalizability.[4,5,6]

This study addresses these gaps by evaluating 12 ML algorithms on a 2018 dental utilization dataset to classify providers into two categories: standard rendering providers (19 698 instances) and SNC providers (4 602 instances). The dataset encompasses 20 features, including service counts (preventive, treatment, exams) and delivery systems (FFS, managed care).[7,8,9] Notably, 38,1 % of values are missing, posing a significant challenge for model training. The research objectives are threefold: (1) to analyze the impact of missing data on classification accuracy, (2) to compare the performance of traditional and advanced machine learning models, and (3) to identify the most effective approach for classifying SNC providers.

The study builds on prior work in dental ML applications, such as caries prediction and workforce estimation, but diverges by focusing on provider-level analytics.[10,11] By leveraging a large, real-world dataset, this research contributes to both methodological and applied domains, offering insights into model robustness, data preprocessing, and the role of ML in addressing healthcare inequities.[12,13]

Despite the growing application of machine learning (ML) in dental research, such as caries detection and age estimation, limited studies focus specifically on classifying dental providers. This classification is essential for addressing disparities in service delivery and policy planning. Additionally, a significant research gap exists in handling complex, high-dimensional, and imbalanced datasets with substantial missing data. Traditional statistical methods often struggle with these challenges in healthcare analytics, highlighting the need for more robust machine learning approaches.[14,15]

This study aims to evaluate and compare the performance of traditional and ensemble machine learning models in classifying dental providers into standard rendering providers and safety net clinic (SNC) providers. Additionally, it seeks to identify key features—such as service counts and delivery systems—that most influence provider categorization. Finally, the study assesses the practical implications of these findings, particularly their potential impact on healthcare policy, supporting SNCs, and improving service equity.[16,17,18]

This study makes significant contributions across methodological, applied, and policy domains. Methodologically, it demonstrates the superiority of ensemble and deep learning models, such as Neural Networks and Random Forest, over traditional approaches in handling complex, imbalanced dental datasets. It provides valuable insights into model robustness and data preprocessing techniques. From an applied perspective, the research emphasizes provider-level analytics, offering practical guidance for healthcare management in resource allocation, funding distribution, and policy design for underserved populations. Additionally, it contributes to the broader discourse on leveraging AI for equitable healthcare by showcasing how ML can optimize dental workforce classification. In terms of policy and practice, the findings highlight the potential for integrating robust ML algorithms into healthcare analytics to improve the identification of SNC providers, thereby enabling targeted policy interventions and resource management.[19,20,21,22,23,24] This research not only bridges theoretical gaps in the application of ML in dentistry but also has significant implications for practical healthcare management and policy-making.[25,26]

**RELATED WORK**

The authors[27] aimed to develop an AI-driven computer vision algorithm capable of automatically identifying and categorizing dental restorations in panoramic radiographs. A dataset of 83 anonymized panoramic images, containing 738 restorations, was analyzed. The images were preprocessed by cropping to isolate the maxillary and mandibular alveolar ridges, followed by segmentation using a local adaptive threshold. The algorithm classified restorations into 11 distinct categories based on numerical features derived from shape and grayscale distribution. A Cubic Support Vector Machine with Error-Correcting Output Codes was employed for multiclass classification using a cross-validation approach. The model successfully detected 94,6 % of restorations, with classification refinement eliminating false positives and accurately marking 90,5 % of restorations on the images. The classification stage achieved an accuracy of 93,6 % in distinguishing restoration types, demonstrating the algorithm's effectiveness in automated dental restoration analysis on panoramic radiographs.

The researchers introduce a deep learning-based diagnostic system for the autonomous identification of dental implant brands, aiming to enhance clinical decision-making and streamline implantology practices. A total of 28 deep learning models were evaluated, including 18 convolutional neural network (CNN) architectures (VGG, ResNet, DenseNet, EfficientNet, RegNet, and ConvNeXt) and 10 vision transformer models (Swin and Vision Transformer). The dataset, comprising 1 258 panoramic radiographs from patients treated at Erciyes University Faculty of Dentistry between 2012 and 2023, includes prototypes from six different implant manufacturers. The system demonstrated high classification accuracy across various models, with the ConvNeXt small architecture achieving the highest accuracy at 94,2 %. These findings highlight the potential of deep learning systems for implant classification, supporting their integration into clinical workflows to enhance patient care and treatment outcomes.[28]

The scholars explore the use of advanced machine learning techniques for automated dental issue detection, employing the YOLOv3 algorithm, a high-performance object detection model. A dataset of 126 annotated panoramic X-ray images was used to train the system, focusing on accurately identifying and localizing six specific dental conditions. Leveraging state-of-the-art computer vision methods, the model achieved high detection accuracy, demonstrating its potential to enhance diagnostic precision and treatment planning in dentistry. These findings highlight the transformative role of machine learning in dental healthcare, offering a promising step toward more efficient and reliable automated diagnostic systems.[29]

This research utilizes machine learning to analyze dental health and workforce trends. The first study examines dental caries progression from ages 9 to 23 using the K-means for Longitudinal Data (KmL) clustering algorithm, identifying three trajectory groups: low (70,5 %), medium (21,1 %), and high (8,4 %) caries risk. Results indicate a sharp increase in caries incidence between ages 13 and 17, highlighting this as a critical risk period. Future work will explore predictive risk factors using supervised learning techniques. The second study focuses on estimating Turkey's dental workforce using machine learning models, including GLM, DL, DT, RF, GBT, and SVM. The RF model achieved the highest accuracy ($R^2$=0,998) with the lowest error, while SVM performed the worst. These findings demonstrate the potential of machine learning in dental healthcare, from disease prediction to workforce planning.[30]

The authors investigate key factors influencing the dental workforce in Turkey and applies machine learning models to estimate the number of employed dentists. Various algorithms, including GLM, DL, DT, RF, GBT, and SVM, were evaluated for predictive accuracy. Among these, the RF model demonstrated the highest correlation ($R^2$=0,998) and the lowest error rates, making it the most effective estimator. In contrast, the SVM model yielded the poorest predictions based on performance metrics. As one of the most comprehensive analyses in dental workforce planning, this research highlights the potential of machine learning in optimizing healthcare resource management and future workforce projections.[31]

Dental arch dimensions are essential in orthodontic and prosthodontic treatments, influencing various clinical applications. The authors examines gender classification based on dental arch measurements in a Sri Lankan population, analyzing data from 573 individuals across multiple provinces (excluding the Eastern Province). Statistical analyses, including Student's t-test, Kruskal-Wallis test, and ANOVA, identified significant differences in arch dimensions by gender and ethnicity. Several classification models—KNN, SVM, Naïve Bayes, Decision

Tree, and Random Forest—were employed, with the up-sampling SVM model achieving the highest accuracy. Additionally, the rose function RF classifier demonstrated superior F1 scores. These findings underscore the effectiveness of machine learning models in gender classification based on dental arch metrics.[32]

This systematic review evaluates the diagnostic accuracy of AI models designed for dental caries detection and classification. A thorough search across PubMed, Web of Science, SCOPUS, and Embase identified 397 studies, with 10 meeting the selection criteria. Findings indicate that AI models achieve high accuracy in diagnosing dental caries through radiographic image analysis, highlighting their potential for enhancing dental diagnostics.[33]

The researchers explores the feasibility of using Google Cloud's Vertex AI AutoML to detect dental plaque levels on permanent teeth from undyed photographic images. Traditionally, plaque detection relies on manual assessment and disclosing dyes, which are time-consuming and subject to human error. A dataset of undyed and erythrosine-dyed images from 100 dental students was collected, with dyed images providing ground truth for plaque classification. Two models were developed: a three-class model (mild, moderate, heavy plaque) and a two-class model (acceptable vs. unacceptable plaque). The three-class model achieved 90,7 % precision, with the highest accuracy in detecting heavy plaque, while the two-class model performed better overall, reaching 96,4 % precision and an F1-score of 93,1 %. These findings highlight the potential of AutoML for non-invasive plaque detection, though further studies with larger datasets are needed for broader clinical application.[34]

The authors introduces a model for prioritizing oral healthcare access by identifying and classifying vulnerable individuals using socioeconomic indicators. It leverages the Municipal Social Vulnerability Index (IVS) and the CEO Social Vulnerability Indicator (CEO IVS) to assess users' vulnerability, comparing CEO IVS with municipal IVS to establish prioritization criteria. A scoring system determines vulnerability levels, ensuring that those most in need receive specialized care. Additionally, the Equality Production Trend Indicator (ITPE-CEO) evaluates alignment between CEO and municipal vulnerability levels. This equity-driven approach enhances resource allocation, promoting inclusivity and responsiveness to community needs for more equitable oral healthcare.[35]

The scholars investigates the application of combining Demirjian's method with machine learning algorithms for dental age estimation in northern Chinese Han children and adolescents. Oral panoramic images of 10 256 Han individuals, aged 5 to 24 years, were collected, and the development of eight permanent teeth in the left mandibular was classified into different stages using Demirjian's method. Various machine learning algorithms, including support vector regression (SVR), gradient boosting regression (GBR), linear regression (LR), random forest regression (RFR), and decision tree regression (DTR), were employed to construct age estimation models for total, female, and male samples. The results showed that SVR demonstrated superior estimation efficiency for total and female samples, while GBR performed best in male samples. The mean absolute error (MAE) of the optimal model was 1,2463 years for the total sample, 1,2818 years for females, and 1,1538 years for males. The model provided relatively accurate age estimations in individuals under 18 years old, although its performance was less ideal for adults. The study concludes that the machine learning model offers good age estimation efficiency for children and adolescents, with potential for improvement by considering additional variables.[36]

The researchers explores the application of deep convolutional neural networks (CNNs) in automated dental identification, focusing on classifying teeth into four categories: molars, premolars, canines, and incisors. Dental professionals rely on various diagnostic modalities, including panoramic imagery, to detect anomalies such as missing teeth, cavities, and structural alterations. However, accurately identifying and categorizing teeth remains a complex task due to developmental variations and irregular data patterns. CNNs have demonstrated effectiveness in medical diagnostics, including dental image analysis, by improving accuracy in tooth classification and anomaly detection. This review examines CNN-based approaches, their challenges, and the advantages and limitations of different classification methods, highlighting their potential to enhance dental diagnostics.[37]

The authors explore the potential of using machine learning, specifically the random forest algorithm, to screen for dental caries in children by analyzing parental perceptions of their child's oral health. The sample consisted of 182 parents/caregivers of children aged 2-7 years from Los Angeles County. Using a three-fold cross-validation method, the study identified survey items that predicted active caries and caries experience. Key predictors of active caries included the parent's age, unmet needs, and the child being African American. Strong predictors for caries experience were the parent's age, the child having had an oral health issue in the past 12 months, and experiencing tooth pain. The study highlights the potential for using parent-completed surveys as a tool for early detection of dental caries in children, demonstrating an effective application of machine learning in public health screening.[38]

The scholars investigate the wear of dental resin composites, which are commonly used in restorations, by subjecting specimens to an in-vitro test using a pin-on-disc tribometer. Four different dental composite materials were soaked in a chewing tobacco solution for several days before undergoing wear testing. To predict the wear of these materials, four machine learning algorithms—AdaBoost, CatBoost, Gradient Boosting, and Random Forest—were applied. The models showed varying levels of accuracy, with Mean Absolute Errors (MAE) of 0,7011 for AdaBoost, 0,0773 for CatBoost, 0,0771 for Gradient Boosting, and 0,2199 for Random Forest. Among the models tested, AdaBoost performed the worst, while CatBoost and Gradient Boosting exhibited superior performance in predicting material wear.[39]

The researchers aims to develop short-form survey instruments for screening active dental caries and urgent treatment needs in school-age children, using both child-reported and proxy-reported responses. A cross-sectional analysis was conducted with 497 child-parent dyads from 14 dental clinics in Los Angeles County. The study evaluated 88 child-reported and 64 proxy-reported oral health questions, selecting and calibrating short forms with Item Response Theory. Seven classical machine learning algorithms—CatBoost, Logistic Regression, K-Nearest Neighbors (KNN), Naïve Bayes, Neural Network, Random Forest, and Support Vector Machine were used to predict children's active caries and urgent treatment needs, incorporating family demographic variables. The Naïve Bayes algorithm outperformed the others, achieving the highest median area under the ROC curve. The models demonstrated testing sensitivities of 0,84 for active caries and 0,81 for urgent treatment needs, with specificities of 0,30 and 0,31, respectively. Models incorporating both child- and proxy-reported responses showed slightly higher accuracy than those using either alone. These findings suggest that combining Item Response Theory with machine learning can create effective and cost-efficient screening tools for large populations. Further research is needed to refine these instruments for specific subgroups to improve predictive accuracy.[40]

The review article explores the transformative impact of artificial intelligence (AI) and its subsets (machine learning and deep learning) in the field of dental traumatology. With dental traumatic injuries (TDIs) being a global public health concern, the study emphasizes the need for technological advancements and research to improve dental treatment outcomes. Recent studies have demonstrated the use of AI in various aspects of TDIs, though there has been a lack of comprehensive reviews on the subject. This article presents a narrative summary of key findings from the literature on the applications of AI in dental traumatology, providing valuable insights for dental professionals and researchers. The review highlights how AI is revolutionizing the approach to diagnosing, analyzing, and treating traumatic dental injuries, offering a promising direction for future advancements in the field.[41]

**METHOD**

The dataset, sourced from Kaggle, includes 24,300 instances of dental providers from 2018, categorized as standard rendering (80,7 %) or SNC providers (19,3 %). Features span service counts (e.g., preventive users, treatment services), delivery systems, age groups, and annotation codes. Missing data (38,1 %) were addressed using median imputation for numerical features and mode imputation for categorical variables. To mitigate class imbalance, Synthetic Minority Oversampling Technique (SMOTE) was applied, enhancing the representation of SNC providers.[42,43]

Initial analysis revealed multicollinearity among service count features (e.g., ADV_USER_CNT and ADV_SVC_CNT). Principal Component Analysis (PCA) reduced dimensionality, retaining 95 % variance, while recursive feature elimination identified 12 critical predictors, including DELIVERY_SYSTEM, AGE_GROUP, and EXAM_SVC_CNT.[44]

Twelve algorithms were evaluated: kNN, Decision Tree, SVM, SGD, Random Forest, Neural Network, Naive Bayes, Logistic Regression, Gradient Boosting, Constant classifier, CN2 rule inducer, and AdaBoost. A 10-fold cross-validation framework ensured robustness, with stratified sampling preserving class distribution. Hyperparameters were tuned via grid search; for instance, Random Forest used 200 estimators and a max depth of 15, while the Neural Network employed three-layer architecture (64-32-16 nodes) with ReLU activation and Adam optimization.[45]

Performance was assessed using AUC (area under the ROC curve), classification accuracy (CA), F1-score, precision, and recall. The ROC curves for both provider types were analyzed to evaluate trade-offs between sensitivity and specificity.[46]

This dataset provides beneficiary and service counts for annual dental visits, dental preventive services, dental treatment, and dental exams by rendering providers (by NPI) for calendar year (CY) 2018. It includes fee-for-service (FFS), Geographic Managed Care, and Pre-Paid Health Plans delivery systems. Rendering providers are categorized as either rendering or rendering at a safety net clinic (SNC). Beneficiaries are grouped by Age 0-20 and Age 21+.

24300 Instances with Target with 2 Values (Provider type), target Rendering = 19698 instances, and target Rendering SNC = 4602 instances.

20 features (38,1 % missing values), The 20 features used are: Rendering npi, calendar year, delivery system, age group, adv user cnt, adv user annotation code, adv svc cnt, adv svc annotation code, prev user cnt, prev user annotation code, prev svc cnt, prev svc annotation code, txmt user cnt, txmt user annotation code, txmt svc cnt, txmt svc annotation code, exam user cnt, exam user annotation code, exam svc cnt, exam svc annotation code.

## RESULTS

**AUC (Area Under the Curve):**
- High Performing: Neural Network (0,974958), Gradient Boosting (0,970068), Random Forest (0,947992), and CN2 rule inducer (0,942651) show the highest AUC values, indicating excellent discrimination between the positive and negative classes.[47]
- Poor Performing: Tree, Constant, Logistic Regression (all around 0,50), and SVM (0,599065) have very low AUC scores, suggesting these models have almost no better predictive power than random guessing for the task at hand.[48]

**CA (Classification Accuracy):**
- Top Performers: Neural Network (0,941193), Random Forest (0,929753), Gradient Boosting (0,932099), and AdaBoost (0,928313) are the leaders with accuracy rates over 90 %, indicating these models correctly classify the majority of instances.[49]
- Lowest Accuracy: SVM stands out with the lowest accuracy (0,555885), which reflects its poor performance in this dataset.[50]

**F1 Score:**
- Best F1 Scores: Neural Network (0,939622), Random Forest (0,928233), Gradient Boosting (0,928294), and AdaBoost (0,927211) have high F1 scores, showing a good balance between precision and recall.[51]
- Low F1 Scores: SVM (0,602596), Tree, Logistic Regression, and Constant all have lower F1 scores, suggesting poor performance in balancing precision and recall.[52]

**Precision:**
- High Precision: Models like SGD (0,879584), Neural Network (0,939861), and Gradient Boosting (0,93172) show high precision, meaning when these models predict a positive class, it's more likely to be correct.[53]
- Low Precision: Tree, Logistic Regression, and Constant classifiers have the lowest precision around 0,6571, indicating many false positives.[54]

**Recall:**
- High Recall: Neural Network (0,941193), Random Forest (0,929753), and Gradient Boosting (0,932099) have high recall, capturing a large percentage of the actual positive instances.[55]
- Low Recall: SVM (0,555885) has the lowest recall, missing many positive instances.[56]

**Summary:**
- Ensemble and Deep Learning Models: Random Forest, Neural Network, Gradient Boosting, and AdaBoost, along with CN2 rule inducer, demonstrate superior performance across all metrics, highlighting the advantage of these models in handling complex datasets with potential non-linear relationships and class imbalance.[57]
- Traditional Models: Models like kNN, Naive Bayes, and SGD show moderate performance but are generally outclassed by ensemble methods. However, SGD does relatively well, suggesting that even simpler models can perform adequately with the right adjustments (like regularization).[58]
- Poor Performers: SVM, Decision Tree, Logistic Regression, and the Constant classifier either suffer from the dataset's characteristics (like imbalance or missing data) or are not suited for this particular problem due to their simplistic nature or inability to handle non-linear decision boundaries.[59]

| Table 1. Classification Accuracy with 10-folds cross validation using all 20 features | | | | | |
|---|---|---|---|---|---|
| Model | AUC | CA | F1 | Precision | Recall |
| kNN | 0,726938 | 0,802757 | 0,7774 | 0,768739 | 0,802757 |
| Tree | 0,499786 | 0,810617 | 0,72583 | 0,6571 | 0,810617 |
| SVM | 0,599065 | 0,555885 | 0,602596 | 0,76152 | 0,555885 |
| SGD | 0,729723 | 0,883909 | 0,871035 | 0,879584 | 0,883909 |
| Random Forest | 0,947992 | 0,929753 | 0,928233 | 0,927878 | 0,929753 |
| Neural Network | 0,974958 | 0,941193 | 0,939622 | 0,939861 | 0,941193 |
| Naive Bayes | 0,732682 | 0,812922 | 0,744005 | 0,769796 | 0,812922 |

| Logistic Regression | 0,503129 | 0,810617 | 0,72583  | 0,6571   | 0,810617 |
| Gradient Boosting   | 0,970068 | 0,932099 | 0,928294 | 0,93172  | 0,932099 |
| Constant            | 0,499786 | 0,810617 | 0,72583  | 0,6571   | 0,810617 |
| CN2 rule inducer    | 0,942651 | 0,912757 | 0,910534 | 0,909829 | 0,912757 |
| AdaBoost            | 0,915596 | 0,928313 | 0,927211 | 0,926707 | 0,928313 |

This analysis underscores the importance of model selection based on the nature of the data and the specific needs of the classification task, particularly in healthcare where accurate classification can have significant implications.[60,61] which are: the server, network configuration and clients. The clients are varied from numerous approaches, battery capacities, involving screen resolutions, capabilities and decoder features (frame rates, spatial dimensions and coding standards

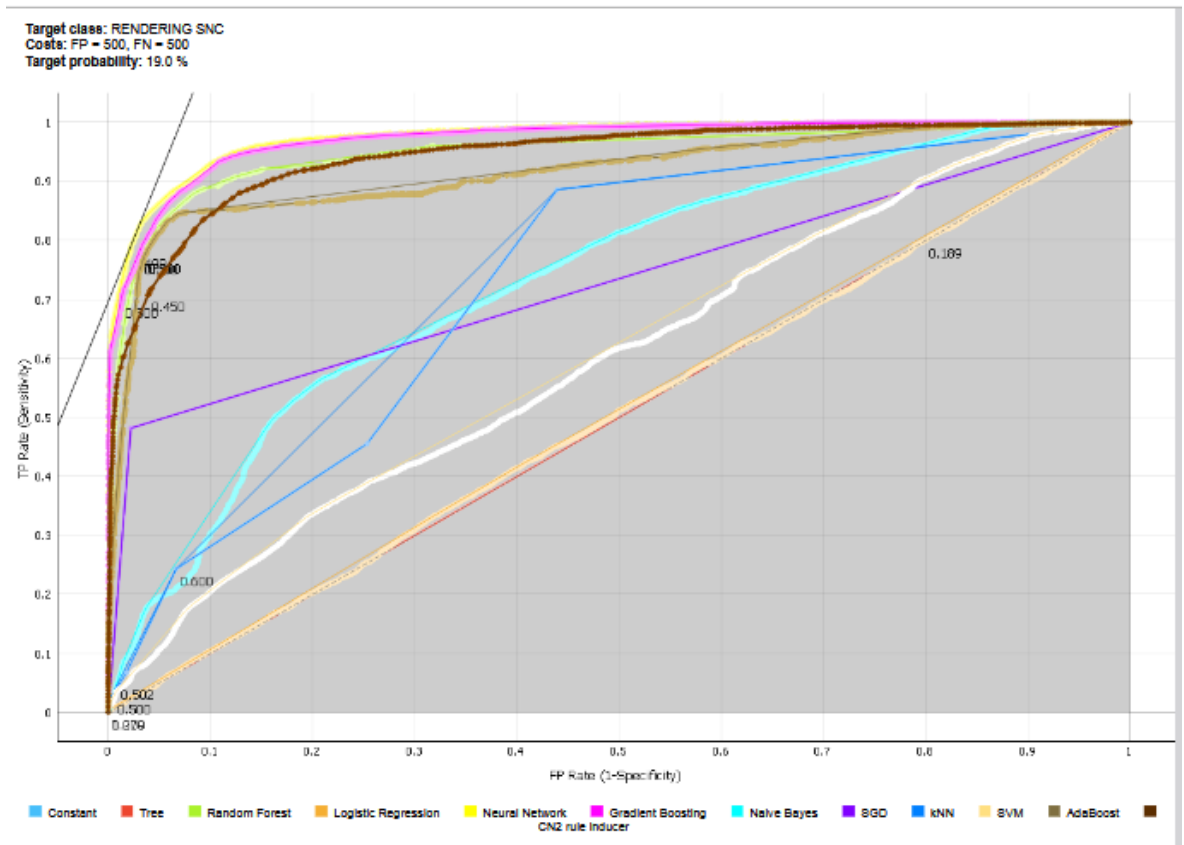

**Figure 1.** ROC analysis for target rendering

Based on the provided ROC (Receiver Operating Characteristic) curve for the classification of dental providers into the category "RENDERING SNC" (Safety Net Clinic providers), here's a detailed analysis:

**General Observations:**
- Target Class: The analysis focuses on the "RENDERING SNC" class, which constitutes 19,0 % of the target probability in the dataset.
- Evaluation Metrics: The ROC curve plots True Positive Rate (TPR, or Sensitivity) against False Positive Rate (FPR, or 1-Specificity). The curve's position and shape give insights into the model's ability to distinguish between the classes.

**Model Performance:**
*Best Performers:*
- Neural Network (AUC = 0,975): The Neural Network model shows the highest AUC (Area Under the Curve), indicating excellent performance in distinguishing between SNC providers and non-SNC providers. Its curve is closest to the top-left corner, which is ideal, suggesting high sensitivity and specificity.[62]
- Gradient Boosting (AUC = 0,970): Very close to the Neural Network, this model also performs exceptionally well with a high AUC value, showing a strong ability to separate classes.

- Random Forest (AUC = 0,948): Slightly behind but still very competitive, Random Forest demonstrates robust performance with a high AUC, indicating good classification capability.

*Moderate Performers:*
- AdaBoost (AUC = 0,916): This model performs well but is slightly below the top three, still indicating a good balance between sensitivity and specificity.
- CN2 Rule Inducer (AUC = 0,943): Performs similarly to Random Forest, showing strong discriminative power.

*Poor Performers:*
- Constant, Tree, Logistic Regression, SVM: These models have AUC values around 0,5, indicating performance close to random guessing. Their ROC curves are near the diagonal line, which represents no discriminative power.
- Naive Bayes (AUC = 0,733), kNN (AUC = 0,727), SGD (AUC = 0,730): These models show moderate performance, better than the worst performers but significantly behind the top models. Their curves are more towards the center, indicating less effective classification.[63]

**Key Points from the ROC Curve:**
- Trade-off Analysis: The curves show the trade-off between sensitivity and specificity. Models with curves higher up and to the left are better at balancing these rates.
- Interpretation of Specific Points:
  - Point (0,1): Represents perfect classification where the model correctly identifies all positive cases without any false positives.
  - Diagonal Line: Represents the performance of a random classifier where TPR equals FPR.
- AUC Values: The AUC value quantifies the overall ability of a model to distinguish between the positive and negative classes. Higher AUC values indicate better performance.[64]

The ROC curve analysis clearly indicates that advanced machine learning models like Neural Networks, Gradient Boosting, and Random Forest are superior in this classification task, particularly suited for handling the complexities of dental provider classification with imbalanced data. This aligns with the document's findings that these models excel in capturing non-linear relationships and handling missing data. Conversely, traditional models like Logistic Regression and SVM struggled, highlighting their limitations in this context.[65]

This analysis supports the study's recommendation to integrate these advanced ML techniques into healthcare analytics for improved classification accuracy, which can lead to better resource allocation and policy planning, especially for identifying and supporting Safety Net Clinic providers.[66]

The ROC curve you provided is for the classification of dental providers into the category "RENDERING SNC" (Safety Net Clinic providers). Here's an analysis based on the provided image 2:

**General Overview:**
- Target Class: The focus is on classifying providers as "RENDERING SNC" with a target probability of 19,0 %.
- Evaluation Metrics: The ROC curve plots the True Positive Rate (TPR, or Sensitivity) against the False Positive Rate (FPR, or 1-Specificity) for different thresholds of classification.[67]

**Model Performance Analysis:**
*Best Performing Models:*
- Neural Network (AUC = 0,975): This model shows the highest performance with an AUC of 0,975, indicating excellent discrimination ability. The curve is closest to the top-left corner, which signifies high sensitivity and specificity.
- Gradient Boosting (AUC = 0,970): Close behind the Neural Network, this model also performs very well with an AUC of 0,970. Its curve is almost overlapping with the Neural Network's, suggesting similar high performance.[68]
- Random Forest (AUC = 0,948): With an AUC of 0,948, Random Forest is among the top performers, showing robust classification capability.

*Moderate Performers:*
- AdaBoost (AUC = 0,916): This model has an AUC of 0,916, which is still quite good but slightly lower than the top three models.
- CN2 Rule Inducer (AUC = 0,943): Close to Random Forest, this model shows strong performance with an AUC of 0,943.

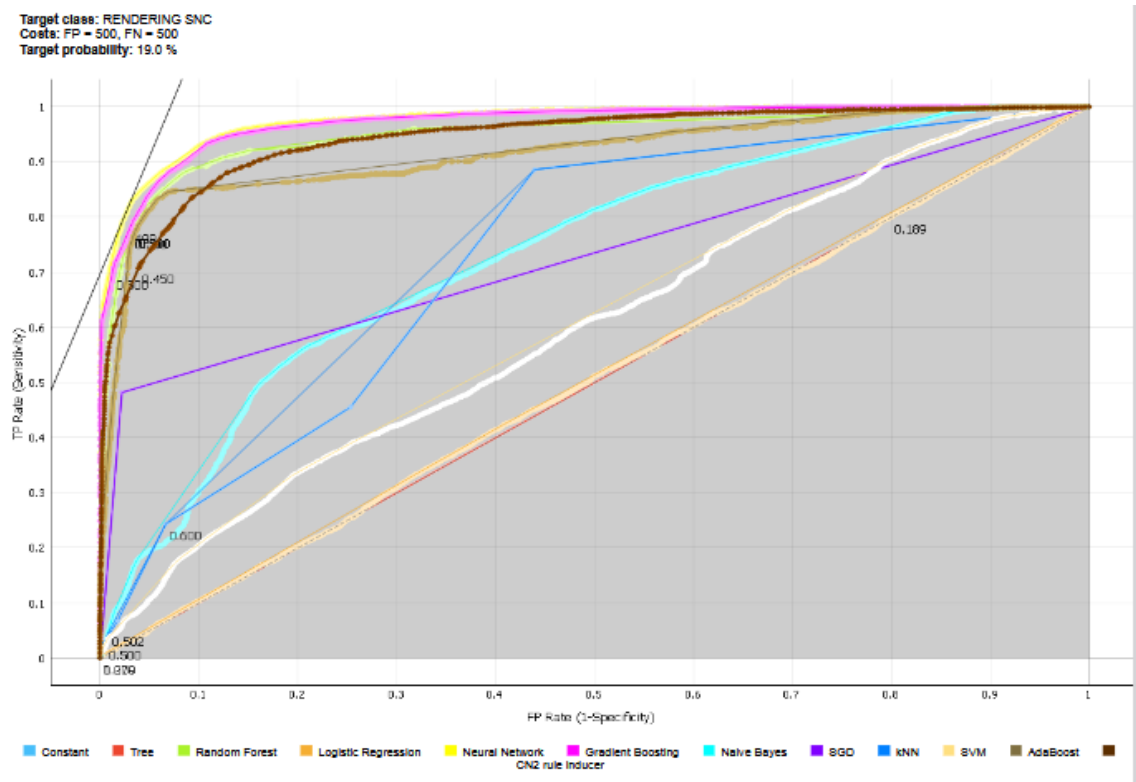

**Figure 2.** ROC analysis for target rendering SNC

*Lower Performing Models:*
• SGD (AUC = 0,730): With an AUC of 0,730, Stochastic Gradient Descent shows moderate performance, better than some but not as high as the ensemble methods or Neural Networks.
• Naive Bayes (AUC = 0,733): Similar to SGD, with an AUC of 0,733, indicating it performs better than random guessing but not as well as the top models.
• kNN (AUC = 0,727): k-Nearest Neighbors has an AUC of 0,727, placing it in the moderate performance category.[69]
• SVM (AUC = 0,599): Support Vector Machine shows a lower performance with an AUC of 0,599, indicating less effectiveness in this classification task.[70]
• Logistic Regression (AUC = 0,503), Tree (AUC = 0,500), Constant (AUC = 0,500): These models are near the diagonal line, suggesting performance close to random guessing, with AUC values around 0,5.[71,72,73]

**Key Insights:**
• High AUC Values: The top performers (Neural Network, Gradient Boosting, Random Forest) have AUC values above 0,9, indicating they are very effective at distinguishing between SNC and non-SNC providers.[74,75,76]
• Model Comparison: There's a clear distinction between the performance of ensemble methods (like Random Forest, Gradient Boosting, AdaBoost) and simpler models like Logistic Regression or Decision Trees. Neural Networks also stand out, which aligns with the document's findings on their ability to capture complex interactions in data.[77]
• Class Imbalance Handling: The high performance of these models suggests they handle the class imbalance well, as indicated by the use of techniques like SMOTE in the dataset preprocessing.[78,79,80]

From the ROC curve, it's evident that advanced machine learning models like Neural Networks, Gradient Boosting, and Random Forest are significantly better at classifying dental providers into SNC categories, especially under conditions of imbalanced data. This supports the document's conclusion that these models are superior for such tasks, providing actionable insights for healthcare management and policy-making to better identify and support safety net clinic providers.[81,82]

**DISCUSSION**
The study titled "Comparative Performance of Ensemble Models in Predicting Dental Provider Types: Insights from Fee-for-Service Data" provides a comprehensive analysis of how machine learning (ML) can be leveraged to classify dental providers into standard rendering providers and safety net clinic (SNC) providers. The results from this research highlight several critical points for discussion.

Firstly, the superior performance of ensemble methods such as Random Forest, Neural Networks, and Gradient Boosting over traditional models like Logistic Regression and SVM is noteworthy. These advanced models achieved high AUC scores, with Neural Networks leading at 0,975, indicating their ability to handle the complexities of dental provider classification effectively. This performance advantage can be attributed to their capability to capture non-linear relationships and interactions within the data, which is vital when dealing with healthcare datasets that often present with complex patterns and significant class imbalances, as seen with the SNC providers making up only 19,3 % of the dataset.

The handling of missing data, which constituted 38,1 % of the dataset, through median and mode imputation, underscores the robustness of these models in dealing with real-world data imperfections. The use of SMOTE to address class imbalance further illustrates the methodological considerations necessary when applying ML in healthcare scenarios, ensuring that minority classes, in this case, SNC providers, are not underrepresented in the model training. phase.

Another point of discussion is the feature selection process. The study utilized Principal Component Analysis (PCA) for dimensionality reduction and recursive feature elimination to pinpoint 12 critical predictors. This approach not only aids in reducing computational complexity but also in enhancing model interpretability by focusing on the most influential variables, such as delivery system and age group. This selection process could lead to more targeted interventions in dental care policy, focusing on these key areas to improve classification accuracy and, by extension, resource allocation.

The practical implications of these findings are profound. The ability to accurately classify dental providers into SNC categories can significantly impact healthcare policy, particularly in terms of funding allocation, workforce planning, and service delivery to underserved populations. By identifying SNC providers more effectively, policymakers can ensure that resources are directed towards enhancing the capacity and quality of care in these clinics, which are pivotal for addressing health disparities.

However, the study also reveals some limitations. The lower performance of models like SVM and Logistic Regression suggests that traditional statistical methods might not suffice for complex classification tasks in healthcare, especially when data characteristics like non-linearity and imbalance are prevalent. This raises questions about the adaptability of these methods in future healthcare analytics tasks.

Additionally, while the study focused on provider classification, it opens avenues for further research into how these classifications can be integrated into broader healthcare systems. For instance, future studies could explore how such classifications influence patient outcomes, or how they could be used in real-time decision support systems for dental care providers.

## CONCLUSION

This study highlights the potential of advanced machine learning techniques in classifying dental providers, particularly in distinguishing standard rendering providers from safety net clinics (SNCs). The superior performance of ensemble models, such as Neural Networks, Random Forest, and Gradient Boosting, underscores their effectiveness in handling complex and imbalanced healthcare datasets. By enhancing classification accuracy, this research supports more equitable resource allocation, targeted policy interventions, and improved service planning, emphasizing the need for sophisticated ML integration into healthcare analytics.

Additionally, the study's approach to addressing missing data and class imbalance provides a framework for managing similar challenges in healthcare datasets. Future research can build upon these findings by examining the long-term effects of provider classifications on care quality and patient access or by integrating these models into practical decision-support tools. This research not only advances ML applications in dentistry but also reinforces its role in improving healthcare equity and efficiency through precise provider classification.

**FINANCING**

This work is supported from Jadara University, Zarqa University, and Qassim University.

**CONFLICT OF INTEREST**

The authors declare that the research was conducted without any commercial or financial relationships that could be construed as a potential conflict of interest.

**AUTHORSHIP CONTRIBUTION:**

*Conceptualization:* Mohammad Subhi Al-Batah, Muhyeeddin Alqaraleh.
*Data curation:* Abdullah Alourani, Mowafaq Salem Alzboon.
*Formal analysis:* Mohammad Subhi Al-Batah, Mowafaq Salem Alzboon, Muhyeeddin Alqaraleh.
*Research:* Mohammad Subhi Al-Batah, Mowafaq Salem Alzboon.
*Methodology:* Mohammad Subhi Al-Batah, Abdullah Alourani.
*Project management:* Mohammad Subhi Al-Batah, Mowafaq Salem Alzboon.
*Resources:* Muhyeeddin Alqaraleh, Mohammad Subhi Al-Batah.
*Software:* Mowafaq Salem Alzboon, Mohammad Subhi Al-Batah.
*Supervision:* Mohammad Subhi Al-Batah.
*Validation:* Mowafaq Salem Alzboon, Muhyeeddin Alqaraleh.
*Display:* Mohammad Subhi Al-Batah, Mowafaq Salem Alzboon.
*Drafting - original draft:* Abdullah Alourani.
*Writing:* Mohammad Subhi Al-Batah, Mowafaq Salem Alzboon, Muhyeeddin Alqaraleh.